\documentclass[final]{cvpr}

\usepackage{times}
\usepackage{epsfig}
\usepackage{graphicx}
\usepackage{amsmath}
\usepackage{amssymb}
\usepackage{graphicx}
\usepackage[frozencache,cachedir=.]{minted}
\usepackage{adjustbox, changepage}

\usepackage{letltxmacro}
\LetLtxMacro\oldttfamily\ttfamily
\DeclareRobustCommand{\ttfamily}{\oldttfamily\csname ttsize\endcsname}
\newcommand{\setttsize}[1]{\def\ttsize{#1}}%
\usepackage{xspace} 
\newcommand{\scenic}{{\sc{Scenic}}\xspace} 

\usepackage[pagebackref=true,breaklinks=true,colorlinks,bookmarks=false]{hyperref}

\def\cvprPaperID{****} 
\def\confYear{CVPR 2021}

\begin{document}
\setttsize{\Large}
\title{{\LARGE{\sc{\texttt{\textbf{Scenic}}}}}: A JAX Library for Computer Vision Research and Beyond}

\author{Mostafa Dehghani \\
Google Brain \\
{\tt\small dehghani@google.com} \\
\rule{2in}{0pt} \and
Alexey Gritsenko\\
Google Brain \\
{\tt\small agritsenko@google.com} \\
\rule{2in}{0pt} \and
Anurag Arnab \\
Google Research \\
{\tt\small aarnab@google.com} \\
\rule{2in}{0pt} \and
Matthias Minderer \\
Google Brain \\
{\tt\small mjlm@google.com} \\
\rule{2in}{0pt} \and
Yi Tay\\
Google Research\\
{\tt\small yitay@google.com} \\
}
\maketitle

\setttsize{\small}

\begin{abstract}
\scenic is an open-source\footnote{\url{https://github.com/google-research/scenic}} JAX library with a focus on Transformer-based models for computer vision research and beyond. The goal of this toolkit is to facilitate rapid experimentation, prototyping, and research of new vision architectures and models. \scenic supports a diverse range of vision tasks (e.g., classification, segmentation, detection) and facilitates working on multi-modal problems, along with GPU/TPU support for multi-host, multi-device large-scale training. \scenic also offers optimized implementations of state-of-the-art research models spanning a wide range of modalities. \scenic has been successfully used for numerous projects and published papers and continues serving as the library of choice for quick prototyping and publication of new research ideas.
\end{abstract}

\section{Introduction}


It is an exciting time for architecture research in computer vision. With new architectures like ViT~\cite{dosovitskiy2020image} taking the world by storm, there exists a clear demand for software and machine learning infrastructure to support easy and extensible neural network architecture research in the field of vision.
As attention models~\cite{dosovitskiy2020image,tay2021omninet,arnab2021vivit} and MLP-only~\cite{tolstikhin2021mlp} architectures become more popular, we expect to see even more research in the coming years pushing the field forward. 

We introduce \scenic, an open-source JAX library for fast and extensible research in vision and beyond. \scenic has been successfully used to develop classification, segmentation, and detection models for images, videos, and other modalities, including multi-modal setups. 

\scenic strives to be a unified, all-in-one codebase for modeling needs, currently offering implementations of state-of-the-art vision models like ViT~\cite{dosovitskiy2020image}, DETR~\cite{carion2020end}, MLP Mixer~\cite{tolstikhin2021mlp}, ResNet~\cite{he2016deep}, and U-Net~\cite{ronneberger2015u}. On top of that, \scenic has been used for numerous Google projects and research papers such as ViViT~\cite{arnab2021vivit}, OmniNet~\cite{tay2021omninet}, TokenLearner~\cite{ryoo2021tokenlearner}, MBT~\cite{nagrani2021attention}, studies on scaling behaviour of various models~\cite{abnar2021exploring,tay2021scale}, and others. We anticipate more research projects to be open-sourced in the \scenic repository in the near future.


\scenic is developed in JAX~\cite{jax2018github} and uses Flax~\cite{flax2020github} as the neural network library, relies on TFDS~\cite{TFDS} and DMVR~\cite{dmvr2020github} for implementing the input pipeline of most of the tasks and dataset, and makes use of for common training loop functionalities offered by CLU~\cite{clu2020github}. 
JAX is an ``ultra-simple to use'' library that enables automatic differentiation of native Python and NumPy functions.
Moreover, it supports multi-host and multi-device training on accelerators including GPUs and TPUs, making it ideal for large-scale machine learning research.

In a nutshell, \scenic is a (i) set of shared light-weight libraries solving commonly encountered tasks when training large-scale (i.e. multi-device, multi-host) models in vision and beyond; and (ii) a number of projects containing fully fleshed out problem-specific training and evaluation loops using these libraries.

\begin{figure*}[t]
    \centering
    \includegraphics[width=1.\textwidth]{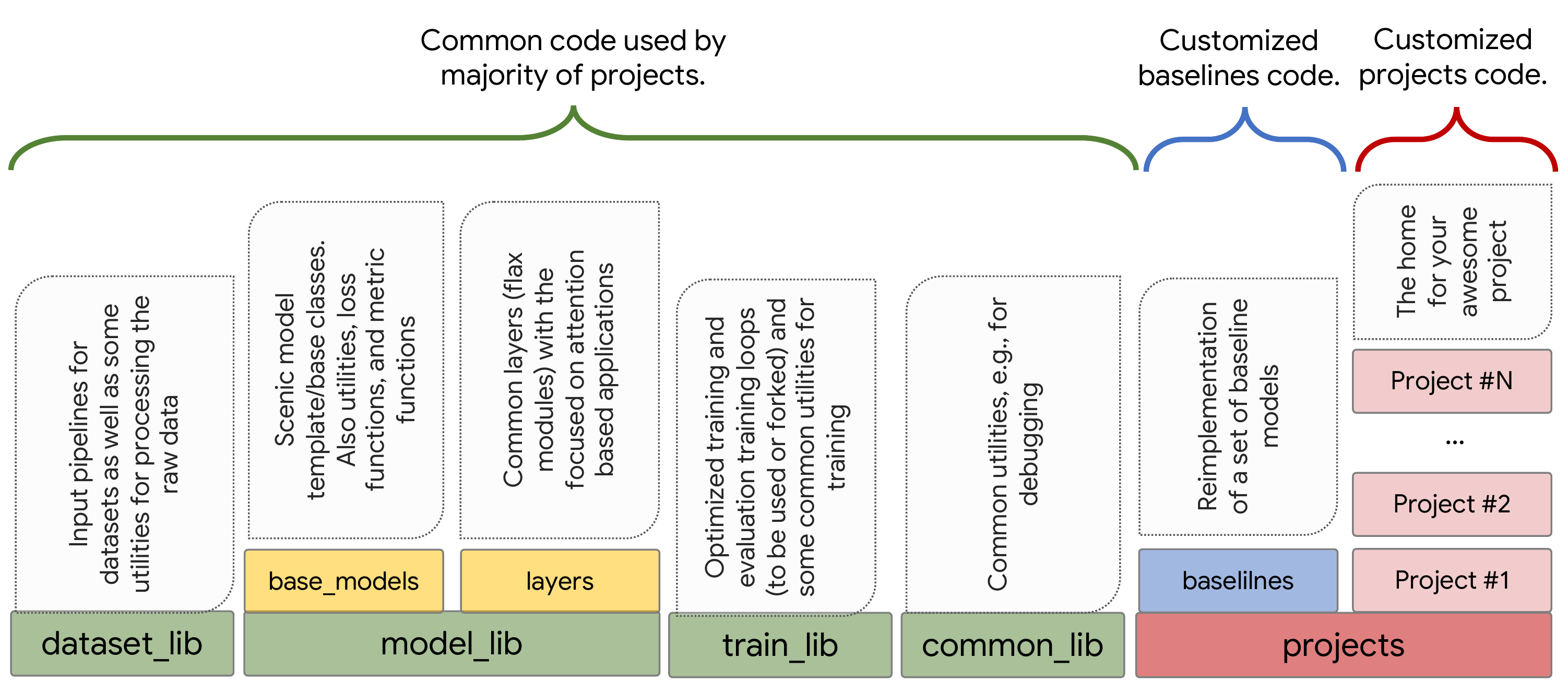}
    \caption{The code in \scenic is organized in \textbf{project-level} part, that is customized code for specific projects or baselines or \textbf{library-level} part, that implements common functionalities and general patterns that are adapted by the majority of projects.}
    \label{fig:scenic_design}
\end{figure*}

\scenic is designed to propose different levels of abstraction. It supports projects from those that only require changing hyper-parameters, to those that need customization on the input pipeline, model architecture, losses and metrics, and the training loop. 
To make this happen, the code in \scenic is organized as either \textbf{project-level} code, which refers to customized code for specific projects or baselines, or \textbf{library-level} code, which refers to common functionalities and general patterns that are adapted by the majority of projects. The project-level code lives in the \texttt{projects} directory.

\paragraph{Philosophy}
\scenic aims to facilitate the rapid prototyping of large-scale models. To keep the code simple to understand and extend, \scenic design prefers forking and copy-pasting over adding complexity or increasing abstraction.
We only upstream functionality to the library-level when it proves to be widely useful across multiple models and tasks.
Minimizing support for various use-cases in the library-level code helps us to avoid accumulating generalizations that result in the code being complex and difficult to understand.
Note that complexity or abstractions of any level can be added to project-level code.

\section{Design}
\scenic offers a unified framework that is sufficiently flexible to support projects in a wide range of needs without having to write complex code.  
\scenic contains optimized implementations of a set of research models operating on a wide range of modalities (video, image, audio, and text), and supports several datasets. This again is made possible by its flexible and low-overhead design. In this section, we go over different parts and discuss the structure that is used to organize projects and library code.

\subsection{Library-level code}
The goal is to keep the library-level code minimal and well-tested and to avoid introducing extra abstractions to support minor use-cases. Shared libraries provided by \scenic are split into:
\begin{itemize}
    \item \texttt{dataset\_lib}: Implements IO pipelines for loading and pre-processing data for common tasks and benchmarks. All pipelines are designed to be scalable and support multi-host and multi-device setups, taking care of dividing data among multiple hosts, incomplete batches, caching, pre-fetching, etc.
    \item \texttt{model\_lib} : Provides several abstract model interfaces (e.g., \texttt{ClassificationModel} or \texttt{SegmentationModel} in \texttt{model\_lib/base\_models}) with task-specific losses and metrics; neural network layers in \texttt{model\_lib/layers}, focusing on efficient implementation of attention and transformer primitives; and finally accelerator-friendly implementations of bipartite matching algorithms~\cite{OTT} in \texttt{model\_lib/matchers}.
    \item \texttt{train\_lib}: Provides tools for constructing training loops and implements several optimized trainers (classification trainer and segmentation trainer) that can be forked for customization.
    \item \texttt{common\_lib}: General utilities, such as logging and debugging modules, and functionalities for processing raw data.
\end{itemize}

\subsection{Project-level code}
\scenic supports the development of customized solutions for specialized tasks and data via the concept of the ``project''. There is no one-fits-all recipe for how much code should be re-used by a project. Projects can consist of only configuration files and use the common models, trainers, tasks/data that live in library-level code, or they can simply fork any of the mentioned functionalities and redefine, layers, losses, metrics, logging methods, tasks, architectures, as well as training and evaluation loops. 
The modularity of library-level code makes it flexible enough to support projects falling anywhere on the ``run-as-is'' to ``fully-customized'' spectrum.

Common baselines such as a ResNet, Vision Transformer (ViT), and DETR are implemented in the \texttt{projects/baselines} project. Forking models in this directory is a good starting point for new projects.

\subsection{\scenic \texttt{BaseModel}}
A solution usually has several parts: data/task pipeline, model architecture, losses and metrics, training and evaluation, etc. Given that much of the research done in \scenic is trying out different architectures, \scenic introduces the concept of a ``model'', to facilitate ``plug-in/plug-out'' experiments. A \scenic model is defined as the network architecture, the losses that are used to update the weights of the network during training, and metrics that are used to evaluate the output of the network. This is implemented as \texttt{BaseModel}.

\texttt{BaseModel} is an abstract class with three members: a \texttt{build\_flax\_model}, a \texttt{loss\_fn}, and a \texttt{get\_metrics\_fn}.

\texttt{build\_flax\_model} function returns a \texttt{flax\_model}. A typical usage pattern is depicted below:

\begin{minted}[
fontsize=\small,
frame=lines,
framesep=2mm,
% baselinestretch=1.1,
% linenos
]{python}
# Get model class:
model_cls = model_lib.models.get_model_cls(
    "fully_connected_classification")
# Build the model, metrics, and losses
model = model_cls(
    config, dataset.meta_data)
# Initialize the model parameters
flax_model = model.build_flax_model
dummy_input = jnp.zeros(
    input_shape, model_input_dtype)
model_state, params = flax_model.init(
    rng, dummy_input, train=False
    ).pop("params")
\end{minted}

And this is how to call the model:
\begin{minted}[
fontsize=\small,
frame=lines,
framesep=2mm,
% baselinestretch=1.1,
% linenos
]{python}
variables = {
# Trainable parameters
"params": params,
# Model state 
# (e.g., batch statistics from BatchNorm)
**model_state
} 
logits, new_model_state = flax_model.apply(
    variables, inputs, ...)
\end{minted}
Abstract classes for defining \scenic models are declared in \texttt{model\_lib/base\_models}. These include the \texttt{BaseModel} that all models inherit from, as well as \texttt{ClassificationModel}, \texttt{MultiLabelClassificationModel}, \texttt{EncoderDecoderModel} and \texttt{SegmentationModel} that respectively define losses and metrics for classification, seq2seq, and segmentation tasks. Depending on its needs, a \scenic project can define new base class or override an existing one for its specific tasks, losses and metrics.

A typical model loss function in \scenic expects predictions and a batch of data:
\begin{minted}[
fontsize=\small,
frame=lines,
framesep=2mm,
% baselinestretch=1.1,
% linenos
]{python}
# Loss function:
loss_fn(
    logits: jnp.ndarray, 
    batch: Dict[str, jnp.ndarray]
    ) -> float 
\end{minted}

Finally, a typical \texttt{get\_metrics\_fn} returns a callable, \texttt{metric\_fn}, that calculates appropriate metrics and returns them as a Python dictionary. The metric function, for each metric, computes $f(x_i, y_i)$ on a mini-batch, where $x_i$ and $y_i$ are inputs and labels of $i$th example, and returns a dictionary from the metric name to a tuple of metric value and the metric normalizer (typically, the number of examples in the mini-batch). It has the API:

\begin{minted}[
fontsize=\small,
frame=lines,
framesep=2mm,
% baselinestretch=1.1,
% linenos
]{python}
# Metric function:
metric_fn(
    logits: jnp.ndarry, 
    label: jnp.ndarry,
    ) -> Dict[str, Tuple[float, int]]
\end{minted}

Given metric and normalizer values collected from examples processed by all devices in all hosts, the model trainer is then responsible for aggregating and computing the normalized metric value for the evaluated examples.

Importantly, while the design pattern above is recommended and has been found to work well for a range of projects, it is not forced, and there is no issue deviating from the above structure within a project.

\section{Conclusion}

Machine Learning (ML) infrastructure is a cornerstone of ML research. Enabling researchers to quickly try out new ideas, and to rapidly scale them up when they show promise, accelerates research. Furthermore, history suggests that methods that leverage the computation available at the time are often the most effective~\cite{sutton1bitter}. 
\scenic embodies our experience of developing the best research infrastructure, and we are excited to share it with the broader community.
We hope to see many more brilliant ideas being developed using \scenic, contributing to the amazing progress made by the ML community for improving lives through AI.

\section*{Acknowledgment}
Scenic has been developed and improved with the help of many amazing contributors. We would like to especially thank Dirk Weissenborn, Andreas Steiner, Marvin Ritter, Aravindh Mahendran, Samira Abnar, Sunayana Rane, Josip Djolonga, Lucas Beyer, Alexander Kolesnikov, Xiaohua Zhai, Rob Romijnders, Rianne van den Berg, Jonathan Heek, Olivier Teboul, Marco Cuturi, Lu Jiang, Mario Lučić, and Neil Houlsby for their direct or indirect contributions to Scenic.

{\small
\bibliographystyle{ieee_fullname}
\bibliography{egbib}
}

\end{document}